# FOCUSING FOR PRONOUN RESOLUTION IN ENGLISH DISCOURSE: AN IMPLEMENTATION

Ebru Ersan and Varol Akman

# BILKENT UNIVERSITY

Department of Computer Engineering
and
Information Science

Technical Report BU-CEIS-94-29

# FOCUSING FOR PRONOUN RESOLUTION IN ENGLISH DISCOURSE: AN IMPLEMENTATION[1]

Ebru Ersan and Varol Akman

September, 1994

[1]Revised version of the first author's M.S. thesis, submitted to the Department of Computer Engineering and Information Science, Bilkent University. Grateful acknowledgment is made to the Scientific and Technical Research Council of Turkey (TÜBİTAK) for a scholarship awarded to the first author during the period of this study. The second author's research is supported in part by a NATO Science for Stability grant (TU-LANGUAGE).
*Address for Correspondence:* Varol Akman, Department of Computer Engineering and Information Science, Bilkent University, Bilkent, Ankara 06533, Turkey. E-mail: akman@cs.bilkent.edu.tr

**Abstract**

Anaphora resolution is one of the most active research areas in natural language processing. This study examines focusing as a tool for the resolution of pronouns which are a kind of anaphora. Focusing is a discourse phenomenon like anaphora. Candy Sidner formalized focusing in her 1979 MIT PhD thesis and devised several algorithms to resolve definite anaphora including pronouns. She presented her theory in a computational framework but did not generally implement the algorithms. Her algorithms related to focusing and pronoun resolution are implemented in this thesis. This implementation provides a better comprehension of the theory both from a conceptual and a computational point of view. The resulting program is tested on different discourse segments, and evaluation and analysis of the experiments are presented together with the statistical results.

# Chapter 1

# Introduction

Anaphora resolution is one of the most active research areas in natural language processing. This study examines focusing as a tool for the resolution of pronouns which are a kind of anaphora. The comprehension of anaphora is an important process. Anaphoric expressions are used and comprehended by humans so often that their importance is usually overlooked. Sometimes it is crucial to resolve an anaphoric expression accurately as the following sentence demonstrates [23, p. 1]:

If the baby does not thrive on raw milk, boil *it*.

There are different kinds of anaphora, pronouns among the most frequently used ones. Sidner formalized the focusing process and devised several algorithms to resolve definite anaphora in English discourse [48]. Her study presents a theory of definite anaphora comprehension. Although the theory draws on a computational framework for the specification of the mechanism of anaphora comprehension, Sidner does not present a full implementation of it.

Focusing is a discourse phenomenon like anaphora. Participants of a coherent discourse center their attention on certain entities during the discourse. Some entities become more salient as the discourse unfolds. These are the entities in focus. Anaphoric expressions are used as a device to refer to these entities. In return, these entities constrain which anaphoric expressions can be used to signal the focus. The entities in the focus space are searched in a predefined order to interpret the anaphoric expressions. The decision is made depending on syntactic, semantic, and inference criteria.

The algorithms of Sidner are implemented in this paper using Lucid Common Lisp [34], and KEE (Knowledge Engineering Environment) [29]. There are algorithms to construct the data structures of the focusing mechanism, a discourse focus, an ordered collection of potential foci, and a stack of old foci. There is a different algorithm for each kind of anaphora in Sidner's thesis but only the ones related to pronoun resolution are implemented in this study. The resulting program is tested on different discourse segments.

While it is almost 15 years old Sidner's thesis is still highly influential. There are some recent studies based on focusing [3, 57]. Our study helps understand the theory of focusing, and its superiorities and deficiencies both from a conceptual and a computational point of view. It can be a first step towards focusing as a tool for resolving pronouns and other kinds of anaphora in Turkish discourse[1].

---

[1]There is not much work done on anaphora in Turkish. Some proposals work on isolated sentences, rather than discourse, using syntactic and surface order analysis [14, 31]. Yet others are on a slightly

In the following chapter, a general review on anaphora is made and different kinds of anaphora are introduced using examples. The third chapter reviews the computational approaches to anaphora resolution. Sidner's study, among the other studies, will be elaborated in this chapter. The issues related to our implementation, and the evaluation and analysis of our experiments are presented in the fourth chapter. The discourse segments used in the experiments can be found in Appendix A.

---

different issue related to anaphora, viz. predicting what can be subject to pronominalization and deletion in the succeeding discourse [11, 13, 15, 30, 54]. Finally, there are some attempts emphasizing the role of context to resolve anaphora within and across sentence boundaries [51, 50, 49].

# Chapter 2

# Anaphora: A General Review

## 2.1 What is Anaphora?

Hirst, who has studied anaphora from a computational perspective, defines it as "the device of making in the discourse an abbreviated reference to some entity (or entities) in the expectation that the receiver will be able to disabbreviate the reference and thereby determine the identity of the entity" [22, p. 4]. The resolution and the generation of anaphora, the former being the disabbreviation of the reference and the latter being the abbreviating reference to an entity according to Hirst's definition, constitute an area of great interest to researchers from different (but related) disciplines such as linguistics and computer science [46]. Because many major problems remain unsolved, this area still actively pursued [28]. In fact, it occupies a central position in the entire field of Natural Language Processing [9].

The entity referred by the anaphoric expression is called the *antecedent*. In the literature, the relation between an anaphoric expression and its antecedent is named as *co-reference*. We illustrate this relation with an example.

> Last night, *John* went to a party. There, *he* met new friends.

In this example, the noun phrase *John* is used to refer to a real (or imaginary) world entity and the anaphoric expression *he* is used to refer to that very same entity. Therefore, they are *co-referential*, i.e., referring to the same entity. The notion of co-reference has some deficiencies. It cannot be used to explain all kinds of anaphora. Its deficiencies will be made clearer by the examples in the sequel, illustrating different kinds of anaphora. The definition of this relation is different in Sidner's case and will be explained in detail in Section 2.3.

Halliday places anaphora in a wider frame [21]. He defines reference as one of the four ways of creating cohesion [37]. He states that there are two types of reference: *exophoric*, referring out of the text to an item in the world (e.g., look at *that*), and *endophoric*, referring to textual items. Endophoric references can be made in three ways: *cataphora*, *anaphora*, and *homophora*. Cataphora are the forward reference tools. For example, in *the house that Jack built*, *the house* refers forward to the specifying *that Jack built*. Anaphora are backward reference tools in Halliday's terminology, as in *Jack built a house. It . . .*, where *it* refers back to *house*. Homophora are self-specifying references to an item of which there can only be one, or only one that makes sense in the context, e.g., *The sun was shining*.

The resolution of anaphora is a complex task because it requires finding the correct antecedent among many possibilities. It involves syntactic, semantic, and pragmatic issues [46]. To see how hard the resolution task can be consider the following example [48]:

John was run over by a truck. When he woke up, *the nurses* were nice to him.

The definite noun phrase *the nurses* is an anaphoric expression, but it does not have an antecedent in the preceding text. Humans resolve this anaphor using *world (extra-linguistic) knowledge*. We know that when people have accidents they are taken to a hospital and that there are nurses in the hospital to take care of injured people. In the example, the definite noun phrase *the nurses* refers to the nurses at the hospital that John has been taken to. An inferencing mechanism is needed to obtain this kind of knowledge. Inferencing, itself, is a complex task and the subject of ongoing research. The following examples, taken from the *Encyclopedic Dictionary of the Sciences of Language* [12], show that there are many dimensions of anaphora:

If *he* comes, *Peter* will be happy.
I ran into *some friends*. These friends (they) spoke to me about you.
I ran into *some friends* who spoke to me about you.
Peter *told me that the weather would be nice.* Jack *too.*
Peter knows my *house*, but not *yours*.

There are different kinds of anaphora such as bound anaphora, one anaphora, sentential anaphora, and pronominal anaphora. Most of the studies were done on pronominal anaphora, especially third person pronouns [46]. There are also studies which try to generalize anaphora [55]. Anaphora, like the other linguistic devices, can be best understood using examples. A simple example of pronominal anaphora can be given as follows:

*John* is a hardworking student. *He* will certainly pass his exams.

In the second sentence, the pronoun *he* is used pragmatically to refer to John. The sentential anaphora can be exemplified by the following pair of sentences [48, p. 66]:

Last week, we went out to the lake near my cottage. *It* was a lot of fun.

Here, *it* is co-referential with the first sentence as a whole, i.e., going out to the lake near my cottage was a lot of fun. The following is an example of bound anaphora [41, p. 430]:

*No one* would put the blame on *himself*.

Bound anaphora (sometimes called bound variable anaphora) occur when the antecedent noun phrases are quantified. The following is a famous example of one anaphora [41, p. 430]:

John lost *a pen* yesterday and Bill found *one* today.

The co-reference relation cannot explain the cases of bound anaphora, and one anaphora. For example, in the last example, *one* (the pen found by Bill) is not co-referential with *a pen* (the pen that John lost). That is, they are probably different pens.

## 2.2 General Approaches to Anaphora Resolution

Efforts towards the resolution of anaphora can be divided into two categories: traditional studies and discourse-oriented studies. The traditional approach has resolution methods on the sentence level [9]; the antecedent of an anaphoric expression is usually searched within the same sentence. These methods generally depend on linguistic knowledge. The possible antecedents are only the set of noun phrases occurring in the preceding text.

The discourse-oriented approach is newer and probably the dominating one. Discourse can be defined informally as a connected piece of text (or spoken language) of more than one sentence (or utterance) [48]. In this approach, researchers try to model the complex structure of discourse. Anaphora, accepted as a discourse phenomena, are tried to be resolved by the help of that structure. Other than the noun phrases occurring in the preceding text, the world knowledge and inferencing are also employed in the resolution process. Furthermore, in addition to the resolution of anaphoric expressions, the discourse-oriented approach considers the generation of appropriate anaphoric expressions. This approach also exhibits some problems but it seems more promising than the traditional approach. Because Sidner's work is one of the leading examples of the discourse-oriented approach, it will be elaborated in the sequel.

## 2.3 Sidner's Approach

Before examining Sidner's work on anaphora resolution, discourse and focusing phenomena should first be studied.

### 2.3.1 Discourse

Sidner takes anaphora as a discourse phenomenon. Therefore, discourse and its structure play an important role in anaphora resolution. Grosz and Sidner recently built a computational theory of discourse structure [19]. It is arguably the most comprehensive theory of discourse, and is widely accepted as a respectable proposal in artificial intelligence (AI). Sidner's work on anaphora resolution came long before the completion of this theory, but it is one of the building blocks of the theory. Therefore, the theory of Grosz and Sidner will be reviewed here.

The theory analyzes the discourse structure. Understanding the discourse structure (i.e., its components and relations among those components) provides a means for describing the processing of utterances in a discourse. According to the theory, any discourse is composed of three separate but interrelated parts. These are the structure of the sequence of utterances called *the linguistic structure*, the structure of purposes called *the intentional structure*, and the state of focus of attention called *the attentional state*.

**Linguistic Structure** Sentences are composed of phrases and phrases are composed of individual words. Likewise, discourses are composed of segments and discourse segments are composed of utterances (i.e., the actual saying or writing of particular sequences of phrases). Finding discourse segment boundaries is an open problem. There are some indicators such as cue-words, and intonation but they do not cover all the cases [9]. Discourse segments are not strictly decompositional. That is, a segment can contain different and nested segments in it. Also, two non-consecutive utterances can belong to the same discourse segment. Discourse segments are important because they effect the interpretation of linguistic devices such as anaphora, just as linguistic devices effect the discourse structure. For example, there are different constraints on the use of certain kinds of referring expressions within segments and at the segment boundaries.

**Intentional Structure** Participants of a discourse have intentions, that is, goals, plans, and purposes. A discourse usually has one general purpose and discourse segments have individual purposes. The relationship among the purposes of the discourse segments gives the structure of discourse. Researchers like Schank [44] and Hobbs [25] try to list and use all of the possible intentions that can appear in any discourse. However, Grosz and Sidner claim that it is not possible to construct a complete list of intentions. Instead, they define two kinds of relations among intentions and use these to reveal the structure of discourse. The two relations are *dominance* and *satisfaction-precedence*. The dominance relation forms a dominance hierarchy of discourse segment purposes. If the satisfaction of a purpose $p$ provides a satisfaction for another purpose $p'$, $p$ is dominated by $p'$. When the order of satisfaction is important, there is the relation of satisfaction-precedence.

**Attentional State** Attentional state is a property of discourse itself. It is an abstraction of the participants' focus of attention. As the discourse unfolds some entities become more salient at some points. These entities are recorded by the attentional state which itself is modeled by a focus space. This is the tool that Sidner provides in her thesis for anaphora resolution [48]. Focusing structure is an important component of the discourse structure. It models the attentional state and coordinates the linguistic and intentional structures as well. Focusing will be elaborated in the upcoming section.

### 2.3.2 Focusing

Focusing, like anaphora, is a discourse phenomenon. Prior to Sidner's study, the concept of focusing was used by Grosz [17]. Grosz also used focusing as the main tool for anaphora resolution. She developed the concept of focus space in her study which later influenced Sidner's work. In the literature, several researchers mention focusing [42] which, after Grosz and Sidner's work, has become a widely accepted tool.

Focus is the thing the communication is about. Although not mentioned explicitly in every sentence, participants of a coherent discourse know that the discourse is about some entity (entities). They should share this knowledge so that they can "catch up" with the information flow as the discourse unfolds. If speakers spell that entity explicitly all over the discourse, it will be infeasible and lead to a dull discourse. Instead speakers use some devices to tell the hearer that one is still talking about that entity. One of these devices is anaphoric expressions.

Focusing is a process. If we can understand its mechanism and find the rules that control this process, then we have an excellent tool for resolving anaphoric expressions. In her thesis, Sidner sketches several algorithms for the focusing process. These algorithms will be presented and explained in the following chapters, but some of the main constructs used by the algorithms necessitate some explanation before examining the use of focusing for anaphora resolution. As an opening, the algorithm "guesses" an *expected focus* at the initial sentence of the discourse. (It is hard to establish the correct focus from the first sentence but something is needed to start on.) If the expected focus is not the correct focus, the algorithm reveals this fact in the consecutive sentence(s) and rejects the expected focus. Of course, the expected focus should be rejected in favor of some other entity. This more favorable entity becomes the *current focus*. Focus is dynamic; it changes during the discourse. There are some supporting constructs such as the actor focus, potential discourse/actor foci, and focus stack. While establishing the current focus, other entities are also traced as possible future foci using these constructs. These notions will be made clearer in the following chapters.

### 2.3.3 Anaphora Resolution

In the literature, anaphoric expressions are usually called referring expressions and a relation of co-reference is defined between an anaphoric expression and its antecedent noun phrase. Sidner claims that noun phrases are not always used to refer and she defines a new relation between anaphoric expressions and their antecedents. The other use of noun phrases is to construct something which can be talked about. Sidner explains such usage with an example [48, p. 16]:

> Mary has a dog.
> He's quite friendly because he wags his tail a lot and wants to play.

In this example, the noun phrase *a dog* is not used to refer, i.e., it does not denote an entity in the world. Its job is to introduce an entity to talk about later. Informally, it is said that *a dog* is the antecedent of *he*. However, because *a dog* is not used to refer to an entity, the relation between them cannot be that of co-reference. Sidner names the relation as *co-specification*. Before defining co-specification, specification should be defined. In her words [48, p. 14] "Specification is the relation between a noun phrase, including its syntactic and semantic interpretation in a sentence, and some database object". She calls a noun phrase and its syntactic and semantic interpretation *the bundle of a noun phrase*. In the theory, whenever a noun phrase is mentioned, it should be thought of as the bundle of a noun phrase. According to the definition of specification, the bundle of the noun phrase *John* specifies a database element which is the representation of the real (or imaginary) world entity John as a person. The anaphoric expression he, if used to talk about John, specifies the same database element. That is, they co-specify. Considering the example above, *he* specifies the same database element as the bundle of noun phrase *a dog*. That is, they both specify the dog that Mary owns; in other words, *he* co-specifies with *a dog*.

Sidner tries to resolve certain kinds of anaphora using the focusing tool. She names the class of anaphora that her theory deals with as *definite anaphora*. This class includes personal pronouns (including possessives) and noun phrases used with a definite article *the*, *this*, or *that*. (She sets the one anaphora aside as a future work.) In her thesis, a different algorithm is sketched for resolving each kind of anaphora.

As mentioned in the preceding section, participants of a discourse center their attention on certain entities as the discourse unfolds. These salient entities form the focus space and anaphoric expressions are usually used to refer to these focused entities. Simply put, anaphora resolution is finding the correct antecedent of an anaphoric expression. It requires searching among a number of noun phrases as possible antecedents, the noun phrases that appear in the discourse, and the noun phrases not mentioned explicitly in the discourse (inferred using world knowledge) as well. When the focusing is used as a tool, the entities in the focus space are searched in an order of preference. This reduces the search space and gives it some order. Because the anaphora are used as a device to refer to the focused entities, it is more feasible (and reasonable) to search the focus space. In the following chapters, once the algorithms for pronoun resolution are given, the use of focusing will become clearer.

The entities in the focus space are possible antecedents. There should be a mechanism to choose among these. The resolution algorithm sends these through a syntactic and semantic filter, and pipes the result to the inference engine for confirmation. (Syntactic and semantic filters will be explained in the following chapter.) The noun phrase which passes these tests is declared to be the antecedent noun phrase. One of the contributions of Sidner is in the inferencing part. Inferencing is an important and necessary mechanism for the resolution of anaphora. The effective application of knowledge for reasoning in AI systems, especially when the number of relevant facts is large, is an important issue and is an area of active study in AI. In one of the most well-known studies, Davis discusses a kind of 'meta-level knowledge' concerned with control of procedure invocation, and illustrates its use in knowledge-based systems [10]. An advantage of focusing is that it controls inferencing. That is, in this theory, the inferencing mechanism does not search for the correct antecedent; its task is simplified to confirmation of an antecedent. (It is still powerful in the sense that it can reject a proposed antecedent if contradictory evidence is found.) In Sidner's thesis, the inferencing mechanism is not defined; only some clues are given. While Sidner admits that it needs a clear definition, she is quick to add that there is ongoing research on the subject and it is not possible to formulate such a definition for the time being.

# Chapter 3

# Computational Approaches to Anaphora

## 3.1 Discourse Understanding

Natural language processing (NLP) is one of the main branches of AI. Natural language is an important property of human intelligence. If we are to build computers which 'simulate' intelligence, they should be capable of communicating with other agents, computer or human, in natural language. Although there are many computational studies on NLP, we are still far from building a general-purpose system. Most of the studies result in systems which can only work in restricted (toy) domains (such as the extraction of certain information from a particular database).

Lately, discourse understanding has become the dominating trend in NLP. To understand a discourse, at least three different kinds of knowledge are necessary [9]. The most obvious one is the linguistic information. It includes the meaning of words, and how grammatically combining these words form sentences that convey meaningful information. Initially, it was thought that the linguistic information was adequate to build a system. However, after a great deal of effort has been spent on research concentrated on linguistic information, it is understood that it was necessary but not sufficient.

The second source of knowledge are the actions performed by the speaker, and the goals and intentions behind them. Linguistic acts comprise speech acts. These include making statements, giving commands, asking questions, and making promises. Speech acts were first studied by Austin [1] and later extended by Searle [45]. For example, a question can be uttered to express a request, as in the sentence "Can you close the door?". Therefore, we need some kind of knowledge to understand the relation between what the speaker uttered and what she actually meant. The success of systems which have the capability of using natural language as humans do is partly dependent on the recognition of intentions of the participants of a discourse.

The third kind of knowledge is found outside the discourse. From the beginning of the computational studies in NLP, it was understood that the information contained in the discourse is not enough to understand a discourse. Background knowledge of the subject of discourse is also necessary. Lack of it makes the discourse incomprehensible or open to a multitude of interpretations. Participants of a discourse usually assume that the other participants have the necessary background knowledge to understand their utterances and that they share a common knowledge of the subject. These assumptions provide them with the ability to make quick references to the objects not explicitly mentioned in the preceeding discourse but existing in the background knowledge of the subject.

## 3.2 Selected Computational Works on Anaphora

### 3.2.1 Early Approaches

Computational studies on discourse processing began in the early 1970s. These concentrated on building computer-based natural language understanding systems. They could achieve tasks such as pronoun understanding and intention recognition only in a limited way. Usually, the systems were designed to work only in certain restricted domains. For example, Charniak built a system whose domain was children's stories [4]. His system tried to find the referents of definite descriptions and pronouns. He encoded the domain information in the form of inference rules and called these rules *demons* [38]. Demons had the ability to decide what to do with a certain piece of information. This system has fundamental difficulties [48]. First of all, even if the domain is restricted, the necessary number of demons to encode the domain information is very large. Another important problem is that since more than one demon can decide to 'fire' at the same time, a control mechanism was needed to handle these situations.

The most famous of the early systems is Winograd's SHRDLU [59]. Its domain was the world of toy blocks. SHRDLU contained a robot arm to move the blocks, a table top, and the toy blocks themselves. It could find the referents of some personal pronouns, and definite descriptions. It could also handle limited versions of one anaphora, elliptical expressions, negation, and quantification. It could learn new word definitions and answer questions about the previous history of the session. Winograd used a well-known heuristic in his system: Pronouns have to agree with their antecedents in person, number, and gender. The antecedent of an anaphor is the last noun phrase that passes a person, number, and gender test [48]. However, as Winograd himself accepts, these rules are not enough for complete anaphora comprehension and in many respects SHRDLU's success was illusionary.

In the late 1970s, studies were carried out on the use of domain knowledge in discourse, and on discourse phenomena other than pronoun understanding. The systems built in this era were still not general-purpose. Some examples are SAM [8], GUS [2], and the Task Dialogue Understanding System [17]. The last one made the distinction among domain knowledge, discourse information, and intention recognition. At the end of this period, it was realized that a complete system cannot be constructed only by putting mechanisms for these three parts together, and that the interaction among these parts is also very important.

### 3.2.2 Discourse Approaches

**Grosz** Grosz built the Task Dialogue Understanding System as a part of her doctoral thesis (reviewed in [9]) and experimentally studied 10 task oriented dialogues and five database dialogues. Her work dealt with the structure of discourse and the notion of focus rather than anaphora resolution, but, of course, a correct formulation of discourse structure and focusing contributes to the comprehension of anaphora. The results of her study show that the structure of the task influences the structure of the dialogues, and additionally sub-dialogues in the task oriented dialogues reflect subparts of the problem to be solved. On the other hand, the database dialogues do not have any complicated global structure.

Grosz developed the concept of a focus space and implemented it using a partitioned semantic network representation of salient objects in a discourse. She resolved non-pronominal definite noun phrases using focus spaces. She gave a noun phrase resolution procedure that can match noun phrases represented by semantic network fragments containing variables to a semantic network database [17]. Focus spaces are used for restricting the search for a match. Using this procedure, non-pronominal noun phrases can be resolved. Grosz's study does not include the resolution of pronouns, and certain cases of the article definite noun phrases. Furthermore, it does not have a complete account for generic/non-generic distinctions. Generics refer to a class while other kinds of definite noun phrases refer to individuals. Sometimes it is hard to distinguish between generic and non-generic usage of noun phrases. The definite noun phrase *the orangutan*, in the following example, can be thought of a generic one unless the sentence is uttered next to an orangutan [48, p. 126]:

> I want to tell you about *the orangutan*.

As mentioned in Chapter 2, Grosz also formulated (with Sidner) a computational theory of discourse structure [19]. The theory handles interesting phenomena such as cue phrases, referring expressions, and interruptions but is difficult to implement. The linguistic structure is partitioned into discourse segments. Although some indicators of segment boundaries are defined, it is not guaranteed that these indicators will always appear. (If this is the case, it is impossible to form the discourse segments.)

**Karttunen** Karttunen inquired how reference by a definite noun phrase or a pronoun is made possible by preceding discourse. He introduced *discourse referents* which are described in Luperfoy's thesis in his own words [36, p. 73]: "In every discourse, there is a basic set of referents which are known to exist although their existence has neither been asserted nor observed during the discourse itself. This set is determined by the common understanding the participants believe they share with regard to their environment." Discourse referents correspond to concepts. They are introduced by either their physical presence or being related to another physically presented entity. Karttunen's rules state that noun phrases and pronouns should refer to a discourse referent. (Only definite noun phrases and pronouns can refer to a discourse referent.)

Discourse referents seem to be the antecedents of Landman's pegs. They can have three sources: the domain (entities which are determined by the shared knowledge of participants), the context (entities which are asserted explicitly), and the discourse (entities which can be inferred from the other existing entities). These sources are approximately the same for Luperfoy's pegs [36] and Webber's discourse entities [58] although they are expressed by these authors in different words. In the following sections, when pegs and discourse entities are introduced, the similarities between them and the features inherited from Karttunen's discourse referents will be apparent.

**Webber** Webber considers discourse as a collection of different types of entities (e.g., individuals, sets, events, actions). *Discourse entities* are the main constructs of her dis-course model. (She calls them "hooks" on which to hang properties.) There exists a discourse entity for each mentioned entity in the discourse model. The speaker assumes that there is a corresponding entity in the hearer's model. Three sources of discourse entities are physical, inferred, and overt mention. A discourse entity (DE) is the referent of an anaphor[1] [48].

In Webber's model there are *invoking descriptions* (ID) of discourse entities. An invoking description is the first mentioning of a discourse entity. In fact, IDs trigger the generation of discourse entities. Antecedents of anaphora are the invoking descriptions. Antecedence relation defined by Webber is similar to Sidner's "co-specification with" relation. Webber made a distinction between her discourse model and a participant's complete memory. Her model is a formal structure validating the sequence of utterances represented as propositions. This distinction is necessary because the participant may not believe what is said during the discourse or may not remember it at all. This distinction is also apparent in Luperfoy's model [36].

Webber investigated how sponsor sentences in a discourse make discourse entities available for potential subsequent anaphoric reference. She gives necessary relations for an ID to be the antecedent of an anaphoric term but does not state which anaphora will be chosen. She uses a typed logic to represent the antecedents of anaphora. Her presentation is able to capture ambiguities similar to the one in the following sentence (both pronouns are possible) [48]:

> Three men who lifted a piano dropped *it/them*.

Webber mainly dealt with one anaphora which includes *one*, *ones*, <null>, *it*, *that*, and *those* according to her classification. She also studied bound anaphora which needs scope identification, and VP-ellipsis and its relation to other anaphoric phenomena. She claims that the accounts of surface structure phenomena and scope of quantification are necessary for the resolution process.

### 3.2.3  Linguistic Approaches

**Partee**  Partee tries to find a uniform treatment of the relation between pronouns and their antecedents. She examines two notions proposed by other researchers, co-reference, and pronouns as bound variables. According to Partee, there are cases in which co-reference fails to be the relation between a pronoun and its antecedent, because the antecedent is non-referential. She claims that in these cases pronouns appear as bound variables. For example, in the sentence below, the co-reference relation does not hold between the indefinite noun phrase *a fish* and *it* [41]:

> John wants to catch *a fish* and eat *it* for supper.

Partee states that in the first part of the sentence a hypothetical world, or a possible state of affairs is described. Fulfillment of this state of affairs introduces a unique object and this object is the actual antecedent of following pronoun. John can eat a fish for supper only if he can catch one. So, the entity referred by the pronoun is not mentioned by a noun phrase in the preceding context. The pronoun has a non-referential indefinite antecedent. (Sidner's definition of co-specification relation can account for this.)

---
[1]This contradicts with Sidner's definition of reference. According to Sidner, referring expressions refer to the entities in the real or imaginary world and not in the discourse representation.

Partee studied one anaphora and bound variable anaphora, and proposed a formalism similar to Webber's. She also pointed out that there is a class of pronouns which can be neither treated as variables nor explained by co-referentiality. These are called *pronouns of laziness*. In the following *it* must be treated as a pronoun of laziness [41]:

> The man who gave *his paycheck* to his wife was wiser than the man who gave *it* to his mistress.

Partee concludes that pronouns must be treated as bound variables but also a criterion must be found to distinguish between pronouns as variables and pronouns of laziness. However, she does not propose one.

**Hobbs** Hobbs' work is a pioneering one in computational linguistics. He developed two approaches for resolving pronoun (including possessives) references [23]. One approach depends purely on syntactic information. The other one is built upon the first one and includes semantic analysis of the text. He, in agreement with Charniak [4], states that once everything else is done, the pronoun resolution occurs as a by-product.

Hobbs developed a naive algorithm for the first approach. It works on the surface parse trees of the sentences in the text. A surface parse tree represents the grammatical structure of a sentence. It includes all the words of a sentence, and syntactically recoverable elements as well. Reading the leaves of the parse tree from left to right forms the original English sentence. The algorithm parses the tree in a pre-defined order and searches for a noun phrase of the correct gender and number. Hobbs also adds some simple selectional restrictions to the algorithm (e.g., dates cannot move). Hobbs tested his algorithm on 100 examples taken from three different sources for the pronouns he, she, it, and they. Although the algorithm is very simple, it was successful in 81.8% of the cases.

In spite of this reasonably successful result, Hobbs states that a semantically based approach should be pursued and cites some reasons. First of all, the naive approach will not yield a total solution. For the examples are taken from written texts, and may not reflect the actual usage of pronouns in English. Also, he claims that, for other reasons, semantic analysis of texts is necessary anyway. Still, he predicts that it will take time before semantically based algorithms perform as accurate as his naive one.

In the second approach, Hobbs describes a system which comprises certain semantic operations to draw inferences from a knowledge base containing world knowledge. There are four basic operations: detecting or verifying the intersentence connectives, predicate interpretation, knitting (i.e., eliminating redundancies), and identifying entities. These operations are designed to recognize the structure and inter-relationships implicit in the text. Pronoun resolution occurs as a by-product of these operations. This approach has some computational problems. The search requires exponential time. But using some techniques, Hobbs claims that the search will be quite fast in 90% of the cases. (However, he does not justify his remarks with empirical results.)

Hobbs also investigated the distance between a pronoun and its antecedent. He defined candidate sets $c_0$, $c_1$, ..., $c_n$ to measure the distance. $c_0$ is the set containing current and previous sentences if pronoun comes before main verb, but only current sentence if pronoun comes after main verb. $c_1$ is the set containing current and previous sentences.

$c_n$ is the set containing current and $n$ previous sentences. The result of the experiments show that the antecedent of a pronoun is found with frequency 90.3% in $c_0$, 7.6% in $c_1$, 1% in $c_2$, 0.6% in $c_3$, and 0.3% in $c_9$ ... The largest distance mentioned is 13.

## 3.3 Sidner's Approach

Sidner's approach has been defined conceptually in Chapter 2. In this section, it will be examined from a computational point of view. Besides exploring the role of focusing in discourse and its relation to anaphora resolution from a theoretical aspect, Sidner devised several algorithms for modeling the focusing process and resolving certain kinds of anaphora. The algorithms will be given in the following chapter.

There are different algorithms for each kind of anaphora, definite noun phrases, third person pronouns in agent position, third person pronouns in non-agent position, third person personal possessive pronouns, *the one ... the other* type anaphora, *this* noun phrases, and *that* noun phrases. NP bundles, i.e., noun phrases including syntactic and semantic interpretation of themselves, are the basic data structures used by the algorithms.

Basic flow in the system is simple. Focusing algorithms find the entities in focus space (entities which are salient in discourse), and the anaphora resolution rules decide the correct antecedent of anaphora under consideration among the entities in focus space. In turn, the results of anaphora resolution rules help the focusing algorithms for the sentence being analyzed. This process is performed for all sentences in the discourse, and for every anaphoric expression in every sentence. Anaphoric expressions are processed in a first-come-first-served manner, from left to right in a sentence. Entities in the focus space are searched in a predefined order. This order is determined according to syntactic and semantic properties of sentences[2]. The entity chosen as a possible antecedent is sent through syntactic and semantic filters, and is forwarded to the inference engine. If it satisfies both of the filters and if the inference mechanism does not find a contradiction, it is chosen to be the antecedent.

Syntactic, semantic, and inference criteria are important because they make the final decision. The syntactic filter includes gender, number, and person agreement, and the disjoint reference computation. The former is, as we have already remarked, a well-known rule used in almost every anaphora resolution system, but the latter needs elaboration. The disjoint reference rule is defined in different ways by various researchers, the most useful one being the definition[3] of Lasnik [32]. Sidner incorporates this rule from the work of Reinhart [43]. It is a syntactic rule and is used for finding the noun phrases which cannot be co-referential within the same sentence. The use of this rule can be understood via the following example of Sidner [47]:

    1a John has the worst luck imaginable.
    1b The professor whose course *he* took gave *him* an F.
    1c The professor who gave *him* an A disqualified *himself* on his   orals.

In (1b), the antecedents of *he* and *him* can be John, or the professor. Both are acceptable syntactically and semantically. Disjoint reference rule tells us that they cannot be co-referential with the professor. In (1c), the rule tells that the antecedent of *himself* is the professor. Semantic criteria includes semantic information about the noun phrases, and rules of scope. For example, the information whether a noun phrase is animate or inanimate can be quite decisive in some cases. There are some problems about the implementation of inference mechanism. Inferencing and suppositions are necessary mechanisms, but Sidner could not define certain kinds of inferences and admitted that making suppositions is an unexplored area of reasoning. Also some suppositions require the modeling of the participants beliefs. However, this subject needs further research to obtain appropriate results. Only the task of inference machine is described. The sentence, with anaphora in it replaced by proposed antecedents, is used by the inference machine to check the consistency with the world knowledge, and the facts stated in the discourse. Inferencing is the only discriminating criterion in some cases. If the anaphora resolution process is examined for the following example, this fact can easily be seen:

> I took my dog to the vet yesterday.
> He bit him in the hand.

---

[2] For example, verb phrases are less frequently focused. Therefore, a verb phrase is not considered as an antecedent until other entities fail to be one.

[3] If NP1 precedes and 'kommands' NP2, and NP2 is not a pronoun, then NP1 and NP2 are disjoint in reference. A 'kommands' B if the minimal cyclic node dominating A also dominates B.

The inference engine should reject *the vet* as the antecedent of *he* using the world knowledge "dogs cannot be beaten in the hand, because they do not have hands" in the previous discourse segment [48, p. 150].

There are some limitations of focusing in anaphora resolution. Focusing cannot help resolve anaphora in parallel structures. We illustrate what is meant by parallel structures by an example [48, p. 179]:

> The green Whitierleaf is most commonly found near the wild rose.
> The wild violet is found near it too.

Here, the pronoun *it* co-specifies with the wild rose. People resolve this kind of anaphora using the parallel structures of the sentences. Focusing would choose Whitierleaf as the co-specification and there are no syntactic/semantic constraints contradicting this choice. Inference engine has also nothing to contradict in this case for two reasons. First, world knowledge is not sufficient for this example. Furthermore, Whitierleaf is an object created by the imagination of Sidner. But this limitation of focusing should not be taken as a deficiency. The uses of parallelism and focusing are fundamentally different. In Sidner's words "The comprehension of definite anaphora which relies on parallelism falls outside of focusing, and some mechanism governing their behavior remains to be discovered" [48, p. 236]. Focusing also does not bring an explanation to certain kinds of pronoun use such as the following [48, p. 176]:

> 1 I went to a concert last night. *They* played Beethoven's ninth.
> 2 John is an orphan. He misses *them* very much.
> 3 I want to meet with Bruce next week. Please arrange *it* for us.

Most of the hearers can easily comprehend the anaphora in 1 and 3 but find 2 a little bit puzzling. Focusing cannot explain this difference. The explanation of how people comprehend *they* as co-specifying with the orchestra in (1) may be that they search the elements associated with the focus (concert in this example), and reach to the conclusion. Another limitation of focusing is that it is sensible only if the sentences have a discourse purpose. The reason is obvious: focusing is valid for coherent discourses in which the participants try to achieve their goals and not to deceive each other [16]. The following discourse segment cannot be said to be coherent if (2b) is uttered after (1) [48, p. 224]:

1 I want to have a meeting with my piano teacher.
2 a Choose the place for me.
  b Eat at the place for me.

In this discourse segment, there is nothing to reject for the syntactic, semantic filters, and the inference machine. One can eat at the place of a meeting. But still (2b) uttered after (1) sounds odd to hearers. Its oddity comes from the bizarre request of eating in this context.

Sidner's work is partly implemented in two systems [48]. One is the Personal Assistant Language Understanding Program (PAL) built at the AI Lab at MIT. The other is the Task Dialogue Understanding System (TDUS) built at the AI Center at SRI International. They are partial implementations because, for example, PAL uses slightly different rules to establish the expected focus. Focus movement is implemented but only the discourse focus and related anaphora resolution rules are incorporated.

## 3.4 Recent Studies

**Luperfoy** Luperfoy built a three-tiered discourse representation [36] and applied it to multimodal human-computer interface dialogues as a part of the Human Interface Tool Suite (HITS) project of the MCC Human Interface Laboratory. Its applications are a knowledge editor for the Cyc Knowledge Base [33], an icon editor for designing display panels for photocopy machines, and an information retrieval tool for preparing multimedia representations. In her system, she deals with context-dependent noun phrases.

The three tiers of the representation are the linguistic tier, the discourse model, and the knowledge base (KB). The first tier is a linguistic analysis of surface forms. A *linguistic object* (LO) is introduced for each linguistic referring expression or non-linguistic communicative gesture performed by each participant of the discourse. In the second tier, an object called a *discourse peg* is created for every concept discussed in the discourse. The third tier, the KB, represents the belief system of an agent participating in the discourse.

The first tier is similar to Grosz and Sidner's linguistic representation [19]. Akin to Sidner's NP bundles, LOs encode both syntactic and semantic analyses of surface forms. However, LOs are anchored to pegs, not to knowledge base objects. Separation of the second and third tiers is very important. This helps represent the distinction between understanding a discourse and believing the information content of it. Sometimes an object in the discourse may be unfamiliar to an agent and she cannot link the corresponding discourse peg to the KB. In this case, an underspecified discourse peg is created and as the discourse unfolds, additional information about that object or inferences add new properties to it and clarify its link to the KB. This ability of late recovery makes Luperfoy's system powerful. Similar to Webber's discourse entities, there are three sources of discourse pegs. They can be introduced by direct mentioning via an LO, as a result of discourse operations performed on one or more existing pegs, and through inferencing. In this system, what is said and what is believed are distinguished; therefore, the KB is not altered automatically during discourse processing. Luperfoy indicates that the information decay is different in each tier. LOs vanish linearly as time passes and as they get older they cannot be linguistic sponsors for anaphora. Discourse pegs decay as a function of attentional focus: as long as the participants pay attention to a peg, it stays near the top of the focus stack and can be a discourse sponsor for referring expressions. Information decay of the KB does not depend on discourse processing. It corresponds to forgetting of stored beliefs.

Luperfoy defines four types of context-dependent NPs [35, p. 4]: "A dependent (anaphoric) LO must be *linguistically sponsored* by another LO in the linguistic tier or *discourse sponsored* by a peg in the discourse model and these two categories are subdivided into *total anaphors* and *partial anaphors*." Total anaphors are co-referential; partial anaphors are not. Definite pronouns are examples of total anaphors. There are different examples of partial anaphors. Some need world knowledge to comprehend the connection between the dependent and the sponsor. An example for this kind of partial anaphors (from Karttunen [27]) is the following:

> I stopped *the car* and when I opened *the hood* I saw that *the radiator* was boiling.

*The hood* and *the radiator* are not mentioned explicitly in the discourse before, but the participants have the world knowledge that cars have hoods and radiators. In some cases generic kinds sponsor indefinite instances [36]:

> Nancy hates *racoons* because *they* ate her corn last year.

Here, *racoons* refers to a class whereas *they* refers to an indefinite instance of that class. The notions of linguistic and discourse sponsoring, and the differential information decay rates of tiers provide semantic interpretation of certain context-dependent noun phrase forms. For example, one anaphora must always have a linguistic sponsor. Another advantage of distinguishing linguistic and discourse levels is that language-specific syntactic or semantic constraints (such as number and gender agreement) are held in the linguistic level and can be overridden by constraints in the higher discourse level.

Luperfoy's work is one of the newest studies on discourse and anaphora resolution. She imports the notion of attentional state (focus) from Grosz and Sidner's theory [19]. Luperfoy tries to resolve one anaphora and deals with generic antecedents which are left as a future work in Sidner's study. Later anchoring of discourse pegs, which can be seen as failure recovery, is another feature of Luperfoy's system which was not available in Sidner's.

**Centering** Centering theory is one of the most recent and influential studies, developed by Grosz, Joshi, and Weinstein [18]. Its computational aspects have foundations in the previously mentioned works of Grosz and Sidner [48, 19]. The centering algorithm was developed by Brennan, Friedman, and Pollard for pronoun resolution [3]. Centering tries to model the process in which the participants of a discourse inform each other about what is the important entity in the discourse. As examined in Chapter 2, one way of doing this is to use pronouns; pronouns refer to the most salient entities.

Centering is a local phenomenon, and works within discourse segments [56]. It does not deal with issues like partitioning discourse into segments, or determining its structure. There are three main structures in the centering algorithm [57]. *Forward-looking Centers* are entities which form a set associated with each utterance. Forward-looking Centers are ranked according to their relative salience and the most ranked entity is called the *Preferred Center*. *Backward-looking Center* is a special member of this set. It is the highest ranked member of Forward-looking Centers of the previous utterance, which is also realized [4] in the current utterance. Using these structures, they define a set of constraints, rules, and transition states (between a pair of utterances). The algorithm incorporates these rules and other linguistic constraints to resolve pronouns.

The most important aspect of centering is the ranking of Forward-looking Centers. The factors determining the ranking are effected by syntax and lexical semantics. The founders of the centering algorithm claim that it works for every language if the correct ranking of Forward-looking Centers is supplied to the algorithm. A computational work of centering in Japanese discourse is presented [57]. Because centering is valid within discourse segments, the antecedents of pronouns are searched within the segment boundaries. If a pronoun occurs in the segment initial utterance, the antecedent is searched in the same utterance.

---

[4] In [57, p. 2], realization is defined as follows: "An utterance U (of some phrase, not necessarily a full clause) *realizes* c if c is an element of the situation described by U, or c is the semantic interpretation of some subpart of U."

# Chapter 4

# The Resolution of Pronouns

## 4.1 Implementation

The algorithms of Sidner should be implemented to examine her work from a computational point of view and to comprehend them truly. The algorithms in her thesis are not detailed enough for a direct mapping to code. That is, one step of an algorithm can hide lots of tiny details. An implementation helps to understand the intricacies of these. Furthermore, and perhaps more importantly, an implementation is needed to experiment with different discourse segments. Our system is implemented with all these issues in mind. It is not an end-product, and surely not recommended to be a part of a commercial natural language processing system. It is just an experimental tool. On the other hand, this does not mean that the program is slow or that one encounters programming errors every time the program is run. It works on every sentence having two noun phrases at most. Confining the number of noun phrases to two is not a severe restriction; the model is valid for most of the well-known discourse segments discussed by several researchers. In the following sections, the data structures used by the program, and the tool with which these structures are realized will be examined. Next the algorithms for focusing and pronoun resolution will be given and their steps will be elucidated.

### 4.1.1 Data Structures

In the implementation of the algorithms, KEE[1] (Knowledge Engineering Environment) [29] is used as a software development tool. KEE is a knowledge system development product that provides software developers with a set of programming tools and techniques for building applications to represent and analyze knowledge.

KEE has a frame system. The basis of this system is a *unit*. Units are similar to the frames of Minsky [38]. Units contain a number of slots. Slots are used to describe the attributes of entities and can hold numerical data, text, tables, graphics, pointers to other units, and procedures written in Lisp. Units can be organized into hierarchies, enabling the knowledge base to be constructed in a more logical manner. Coupled with KEE's inheritance mechanism, this allows for efficient storage and reasoning. Also, KEE provides a user interface and related tools which help the programmer during the construction phase and afterwards for building her own user interface.

---

[1] KEE (Knowledge Engineering Environment) is a trademark of IntelliCorp, Inc.

The data structures used by the program are represented in the form of units. Sentences and their constituent words are represented by units. There is one class of unit for each type. Sentences are represented by the class *sentence*, noun phrases by *np*, and verb phrases by *vp*. Because they share some common properties (e.g., belonging to a sentence), *np* and *vp* are constructed as subclasses of the class *word* using the inheritance mechanism of KEE. Each discourse segment is saved into a separate knowledge base containing these main classes. Sentences and words of discourse segments are represented by the instantiations of these classes. As mentioned above, Sidner's algorithms require the semantic and syntactic analyses of the sentences and words to operate on. We do not currently have a tool to make this kind of analysis automatically. Therefore, the data structures are constructed manually. This only increases the time to construct the discourse segment on the computer.

The slots of the units contain the results of the semantic and syntactic analyses from which the algorithms benefit. (A few slots are used only for programming purposes, e.g., to hold a flag.) There are 13 slots in *sentence*. One of the slots indicates the type of the sentence; the possibilities are is-a (e.g., John is the president), there-insertion (e.g., There was a house which looked horrifying), cleft (e.g., It was Mary who knocked at the door), pseudo-cleft (e.g., The person who knocked at the door was Mary), and normal (ordinary sentences). The type of a sentence bears some information about the focus. The cleft, pseudo-cleft, and there-insertion sentences syntactically mark a special object, most probably the focused item. Two other slots hold the cleft, and pseudo-cleft items (Mary, for the above examples) if the sentence is of that type. Two slots point to the previous and consecutive sentences. These slots are necessary to move back and forth in discourse. Four slots point to the subject (*np* unit), verb phrase (*vp* unit), and two noun phrases (*np* units) of the sentence. The subject and noun phrases can be a collection of several nouns (e.g., John, Mary, and their dog went to the school together). There are examples of this type examined in the following sections. One slot holds the position of anaphoric expressions such as subject, first noun phrase (np1), and second noun phrase (np2). The last slot is used to indicate whether the sentence contains do-anaphora or not. (Do-anaphora is explained in Section 4.1.2.)

There are 11 slots in *np*. If the noun phrase is anaphoric, one slot is used to discriminate the kind of anaphoric expression, viz. definite noun phrase or pronoun. As a result of the program, the co-specification of an anaphoric noun phrase is written to a slot when found and read from this slot when necessary. One of the slots holds the information whether the noun phrase implicitly specifies another noun phrase or not. This information is used by the focusing algorithm. The sentence that noun phrase belongs to and the position of it in the sentence are pointed by two slots. Four slots hold the gender, number, person, and life-form of the noun phrase. Gender slot can have the values masculine, feminine, neuter, or any combination of these. (For example, the gender of the pronoun *I* can be masculine or feminine.) The value of the number slot is either singular or plural. Person slot has the values first singular, second singular, etc. Life-form of a noun phrase can be animate, inanimate, or none (if it is not known)[2]. Obviously, one slot holds the spelling of the noun phrase. (For example, that slot of a unit representing the noun phrase *kite* has the value "kite".)

*vp* shares two of its slots with *np*; these are the slot which points to the sentence it belongs to, and the slot that holds the spelling of the verb phrase. It has one additional slot which holds the theme of a verb phrase. The theme concept has been studied by many researchers and not only for resolving anaphora [20, 26]. A complete account of thematic relations has not been obtained yet. Theme is useful for finding the expected focus. Expected focus is likely to be the object of an action. However, this explanation does not yield a solution for all kinds of sentences. The theme concept can be the answer. Consider the following examples [48, p. 63]:

> 1 Mortimer sold *the book* for 10 cents.
> 2 *The chest* is standing in the corner.
> 3 Please focus on *the star of India* in the case on the left.

---

[2]For example, the pronoun *they* can be used to refer to animate or inanimate objects, so the life-form slot of a unit which represents the noun phrase *they* should have the value 'unknown'.

The theme of the verb is *the book* for the first sentence, *the chest* for the second, and *the star of India* for the third. Sidner claims that the theme of a verb is the direct object in most of the cases. This is true for the first and second sentences, but not for the third. Because of the preposition *on*, *the star of India* cannot be marked as the direct object, and hence the theme. The definition of theme should include some semantic foundations besides the syntactic ones. Sidner defines theme as the verb relation that indicates the property of being affected by the action of the verb. This is a useful heuristic, good for focusing purposes. For the affected object is likely to be the focus of further communication.

### 4.1.2 Algorithms

The algorithms presented in this section are slightly modified versions of Sidner's algorithms. Some of the modifications are made to adapt these algorithms to our system. Some of the modifications are made to correct some failures observed during the experimentation phase. (When the algorithms are examined individually, details of the modifications will be explained.) The programming language in which the algorithms are coded is Lisp [34] which is fully supported by KEE.

#### Focusing Algorithms

When the program is run, it needs something to start on. It is hard to establish (even for human beings) the right focus from the first sentence of a discourse. A special routine called *The Expected Focus Algorithm* is used for "guessing" the focus from the initial sentence. There are some syntactic and semantic clues marking the focus. The type of a sentence is among these. Is-a or there-insertion type of sentences are used to signal out one object which is the most possible candidate of discourse focus. The thematic relations of a verb are other possible candidates. The preference among the thematic relations cannot be easily found. Because the algorithm needs an order to choose among these relations, Sidner uses a simple scheme of sentence surface order.

**The Expected Focus Algorithm**
The expected focus is chosen as:

1. The subject of a sentence, if the sentence is an *is-a* or a *there-insertion* sentence.
   This step presumes information from a parse tree what the subject and the verb are and whether the sentence is of there-insertion type.

2. Otherwise, the first element of the default expected focus list (DEF list), computed from the thematic relations of the verb, as follows:
   Order the set of phrases in the sentence using the following preference schema:
   – theme
   – all other thematic positions except the agent
   – the agent
   – the verb phrase
   This step requires a list of the surface order of the noun phrases, and a data structure which indicates which noun phrases fill which thematic slots in the verb.

In the flow of the program, pronoun rules are applied after the expected focus algorithm. Because pronoun resolution rules operate on entities created by the focusing algorithm, the focusing algorithm will be examined first. It consists of eight steps. Step 1 is designed for the case where the sentence contains do-so, do-it type anaphora [48, p. 61]:

   Ben ran. To do so, Ben had put on his new tennis shoes.

After the first sentence is processed, the expected focus is Ben. The second sentence contains a do-so type anaphor which co-specifies with the verb of the previous sentence. Therefore, the current focus is set to the verb phrase which is the last member of alternative focus list. In Step 2, the special case of focus sets is handled. To detect such a condition, the algorithm checks the current focus (CF) to see whether it is nil or not. CF can be nil if the sentence does not contain any anaphoric expressions to confirm or reject the focus. After the expected focus is found, CF is established in the light of the anaphoric expressions. If there is no anaphoric expression, it is set to nil. This is a very special odd case [47]. The existence of focus sets is one of these special cases. Consider the following example [48, p. 76]:

   John and Mary sat on the sofa and played cards.
   Henry read a book.
   At 10 p.m. they went to Joey's Bar to hear a new rock group.

It is seen that the focus is not the sofa, cards or book. The discourse is about John, Mary, and Henry, and what they did. The focus should be collected over several sentences. If there is an indication of focus sets (CF becoming nil), three focus sets are collected in this step. These are theme, actor, and verb phrase sets. The theme set contains cards and book, the actor set contains John, Mary, and Henry, and the verb phrase set contains the actions of sitting, playing, and reading. This process is continued until an anaphoric expression is encountered in the consecutive sentences. Step 3 is self-explanatory. It exhibits some important focusing features. We see that the entities in the non-agent position are preferred over the entities in the agent position. Also, pronouns rather than definite noun phrases are used for marking the focussed entity. Therefore, their co-specifications are preferred as focus. Step 4 is the simplest case, the focus is still the same entity. In Step 5, the sentence surface order used by the expected focus algorithm is utilized. The choice is made according to that order. In Step 6, another important feature of focusing is presented, the use of *focus stack*. Sometimes discourse returns to an entity previously under discussion. This process is called *focus popping*. A simple last-in first-out stack is suitable to model the popping computationally. The discourse in the following example exhibits this feature [48, p. 88]:

> Wilbur is a fine scientists and a thoughtful guy.
> He gave me a book a while back which I really liked.
> It was on relativity theory.
> It talks about quarks.
> They are hard to imagine,
> because they indicate the need for elementary field theories of a complex nature.
> These theories are absolutely essential to all relativity research.
> Anyway, I got it while I was working on the initial part of my research.
> He's really a helpful colleague to have thought of giving it to me.

In this discourse, the focus moves from Wilbur to the book, to relativity theory, to quarks, and to elementary field theory. Then it pops back to the book. After this pop is made, Wilbur can be co-specified with the pronoun *he* easily. The intervening foci (relativity theory, quarks, and elementary field theory) are discarded. This is the same as the behavior of a stack. Usually, the pop-back occurs by the use of a definite noun phrase because it has the same head as the stacked entity and this makes the recognition easier. Pronouns can also be used but there are some restrictions on their use. The pronoun that co-specifies with a stacked entity should not be acceptable as co-specifying with discourse focus, or potential foci. (In such a case, it may be impossible to resolve the pronoun both for humans and computers.) In Step 7, a case related to definite noun phrases is handled. Because this work excludes the resolution of definite noun phrases, their co-specifications and other relevant information is given to the system manually. In the following, the implicit specification is exemplified:

> Yesterday I was driving *my car* on the highway.
> *The engine* suddenly stopped.
> I took *it* to the mechanic.

The definite noun phrase *the engine* implicitly specifies an element associated with the discourse focus *my car* in this example. In the third sentence, the co-specification of *it* is *my car*. It is observed that the focus has not changed from first sentence to the third because *the engine* implicitly specifies the current focus.

In the last step (Step 8) of the algorithm, the actor focus is established[3]. Actors are distinguished from the discourse focus because in many discourses, the actors are mentioned and pronominalized besides the discourse focus. There are potential actor foci just as potential foci[4]. They can be the antecedents of pronouns in agent position. The potential actor focus list is simply constructed by appending the animate objects which are not in agent position.

**The Focusing Algorithm**

There are several main data structures manipulated by the focusing algorithm. When the discourse is initial, CF (current focus) is set to the expected focus and the ALFL (alternative focus list) is set to the DEF (default expected foci). When the discourse is in progress, CF is set to the discourse focus and ALFL is set to the PFL (potential focus list). Also, an empty focus stack is available globally prior to the first run. The following steps are used to retain the current focus or reject it in favor of another entity. When the CF is moved to another entity, it is stacked in the focus stack.

---

[3] "An actor is an animate object which may function as the agent of a particular verb" [48, p. 152].

[4] "A potential actor is a noun phrase which specifies an animate element of the database and does not occur in agent position" [48, p. 153].

1. **Do-anaphora:** If the sentence contains do-anaphora, take the last member of the ALFL as the focus. Ignore steps 2 through 6.

2. **Focus Set Collection:** If there is no CF[5], there is an occurrence of focus sets. When no definite anaphora have appeared in the current sentence, continue collecting focus sets. If an anaphor appeared and it is not in agent position, take its co-specification as focus.

3. **Choosing Between CF and ALFL:** If there are anaphora which co-specify both the CF and some member of ALFL, take as focus whichever is not in agent position. If both are non-agents, retain the CF as focus unless only the ALFL member is mentioned by a pronoun. In that case, move the focus to the ALFL member.

4. **Retaining the CF as Focus:** If there are anaphora which co-specify only the CF, retain the CF as focus.

5. **ALFL as Focus:** If the anaphora only co-specify a member of ALFL, move the focus to it. If several members of the ALFL are co-specified, choose the focus in the manner suggested by the expected focus algorithm.

6. **Focus Stack Use:** If the anaphora only co-specify a member of the focus stack, move the focus to the stack member by popping the stack.

7. **Implicit Specification:** If a definite noun phrase implicitly specifies an element associated with the focus, retain the CF. If specification is associated with a member of ALFL, move focus to that member. In both cases, flag the definite noun phrase as implicit specification.

8. **Actor Focus:** The Actor focus (AF) is the agent in the current sentence (and its specification), if one exists; otherwise, the actor focus remains unchanged. If the actor focus takes on a new specification, the old actor focus is stacked in the actor focus stack.

A potential focus list (PFL) is constructed for every sentence. It is determined according to the following algorithm:

1. If a cleft or pseudo-cleft sentence is used, the potential focus is the cleft item if and only if the entity in the non-clefting position co-specifies the focus. (When it does not, the sentence is incoherent.)

2. Otherwise, order a potential focus list of all the noun phrases filling a thematic relation in the sentence, excluding the noun phrase in agent position and the noun phrase which co-specifies the focus if one exists. The last member of the PFL is the verb phrase of the sentence.

---

[5] When there are not any definite anaphora in the consecutive sentences to confirm (or to reject) the CF, CF becomes nil [47].

**Pronoun Resolution Algorithms**

The rules for first and second person pronouns are simple. The first and second person singular pronouns co-specify with the speaker and hearer of the discourse. There is one exception to this rule, the case in which the first and second person pronouns appear in quotes. For the first and second person plural pronouns, first the actor focus, then the focus for a class with the speaker (hearer) in it are checked. *I* and *you* are deictic pronouns. Sidner states that her rules do not govern them accurately though some of the same techniques can be used for setting a deictic focus in the first utterance and then updating it according to the change of speaker [47]. Our program simply marks the co-specifications of *I* and *you* and their variants as speaker or hearer.

Third person pronoun rules are presented in the form of flowcharts in Figures 4.1–4.3. These rules are divided into two subparts. One part includes the rules for the third person pronouns in agent position, and the other includes the rules for the ones in non-agent position. There are some abbreviations in the figures. DF stands for discourse focus, AF for actor focus, PDF for potential discourse foci, and PAF for potential actor foci. (For the sake of brevity, these abbreviations will also be used in the sequel.) In the flowchart, there are some endpoints which need explanation. One endpoint is marked SUCCESS and indicates that the antecedent of the pronoun under consideration is found. The question "Is ... acceptable as co-spec?" should be answered in order to be able to reach such a conclusion. This check is made using the syntactic and semantic filters explained in Section 2.3. These filters look for the gender, number, person, and life-form agreement. Also, the disjoint reference rule is applied to find the entities which cannot be co-specifying. As explained in Section 2.3, Sidner states that, once these checks are made, an inference mechanism should be used to confirm the choice. However, she does not give a definition of the inference mechanism. Therefore, an inference mechanism is not implemented in our program. In the following sections, when the examples are examined, the effect of inferencing on the results will be presented.

Some of the endpoints mark conditions other than SUCCESS that can be met during the resolution process. The rules do not explain how to proceed from these conditions because these conditions cannot be explained by focusing. Nevertheless, they can at least be marked when encountered. There are four such conditions. *Backwards non-antecedent pronoun condition* occurs when there is no antecedent of the pronoun in the preceding text [48, p. 176]:

> I saw Mr. Smith the other day; you know, *she* died last year.

This example may sound odd to many hearers. It can be made understandable if the hearer is informed that Mr. Smith had a wife. The pronoun *she* does not have an antecedent in the discourse, it is a backward non-antecedent pronoun. There is a similar condition called *backwards non-antecedent pronoun condition or forward co-specification*. In this case, the discrimination between the two conditions cannot be made. The forward co-specification condition can be met in the case of pronouns in non-agent position. When the pronoun is uttered before its antecedent, this condition is met:

> I saw *it* all of a sudden. A large grey snake was resting on the floor.

Here, *it* co-specifies with the snake which is uttered after the pronoun. The focusing is not able to explain this usage, i.e., cataphora [5]. The third condition is *potential actor ambiguity condition* [48, p. 155]:

*Potential Actor Ambiguity Condition*
Whenever a pronoun may co-specify the actor focus and a single potential actor exists, expect a possible ambiguity. To resolve, (1) look for evidence supporting the statement in which the pronoun occurs, evidence which is true of the actor focus as the co-specification, but not of the potential actor. If this is found, the actor focus is the co-specification. (2) However, if there is evidence true for both, choose the actor focus but indicate ambiguity. (3) Choose the potential actor when evidence exists for it but not for the focus.

She proposes three steps to resolve this condition but these are too vague to be implemented as she herself accepts[6]. The last condition is *unreliable pronoun use*. This condition is met when the actor focus is the same type of pronoun as the pronoun under consideration but cannot be accepted as its co-specification.

The rules for the third person pronouns in agent position (cf. Figures 3.1–3.2) start with questioning the existence of AF and DF. If any of them is missing, there must be focus sets constructed by the focusing algorithm. If this is not the case, or the animate focus set (animate because the antecedent of the pronoun in "agent" position is searched) is not acceptable as co-specification, this means that *backwards non-antecedent pronoun condition* is met. Step 2a states the well-known *recency rule* [48, p. 144]:

*Recency Rule*
If the pronoun under consideration occurs in the subject position, and there is an alternate focus list noun phrase which occurs as the last constituent in the previous sentence, test that alternate focus list phrase before testing the current focus. If acceptable both syntactically and inferentially, choose the alternate focus list phrase as the co-specification of the pronoun.

This is one of the oldest anaphora resolution heuristics. It rules out the focusing but Sidner claims that she observed this rule to be consistently accurate. In the following sections, effects of this rule on the result of resolution process will be explained on some examples. Step 3 performs the rule called *Animate Discourse Focus Rule* [48, p.152]:

*Animate Discourse Focus Rule*
A pronominal expression in agent position which meets person, number, and gender agreement with the actor focus co-specifies with the actor focus unless the pronoun may also co-specify the discourse focus. In the latter case, if the discourse focus was established in a sentence before the actor focus, and meets person, number, and gender agreement, the discourse focus is the intended co-specification.

This rule is utilized in situations where the discourse focus is animate in addition to the actor focus. The rule does not state what to do when both foci meet person, number, and gender agreement and they are established at the same sentence. Additional information is necessary to resolve this ambiguity. In Step 4b, *the potential actor ambiguity condition*, explained above, is checked. Step 5 asks whether the AF is a pronoun of same type as the pronoun under consideration. This is a frequently observed case. If the actor focus is pronominalized, it is likely to be pronominalized in the same way in the following sentences, if it is still the focus. In Steps 6b-9, according to the number of the pronoun under consideration, AF, PAF, and AF stack are checked in the order stated in the flowchart. In the last steps, discourse focus and potential discourse foci are checked in turn. If none of them is acceptable as co-specification, *backwards non-antecedent pronoun condition* holds.

---

[6]While experimenting, we have encountered some examples meeting this condition. We observed that we could resolve in some of the cases if we send the actor and potential actor foci to the syntactic and semantic filter. If both are acceptable, nothing further is to be done. If only one of them can pass through the filters the algorithm proceeds and reaches to the correct antecedent. We discussed this with Sidner [47] and she has found it reasonable but also indicated that this cannot be helpful in all of the cases. The examples of this condition can be found in the following sections.

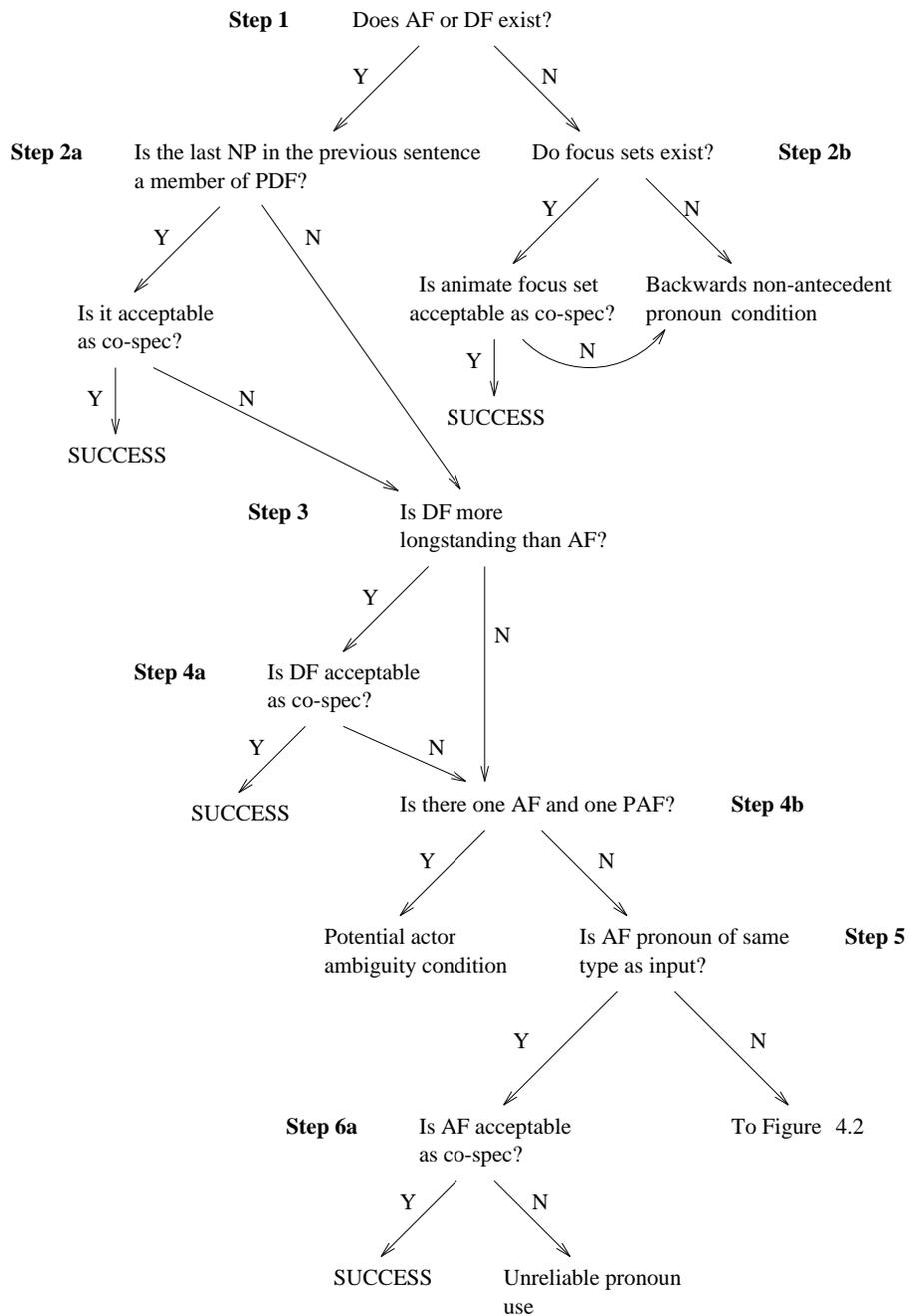

Figure 4.1: Flowchart for the third person pronoun in agent position.

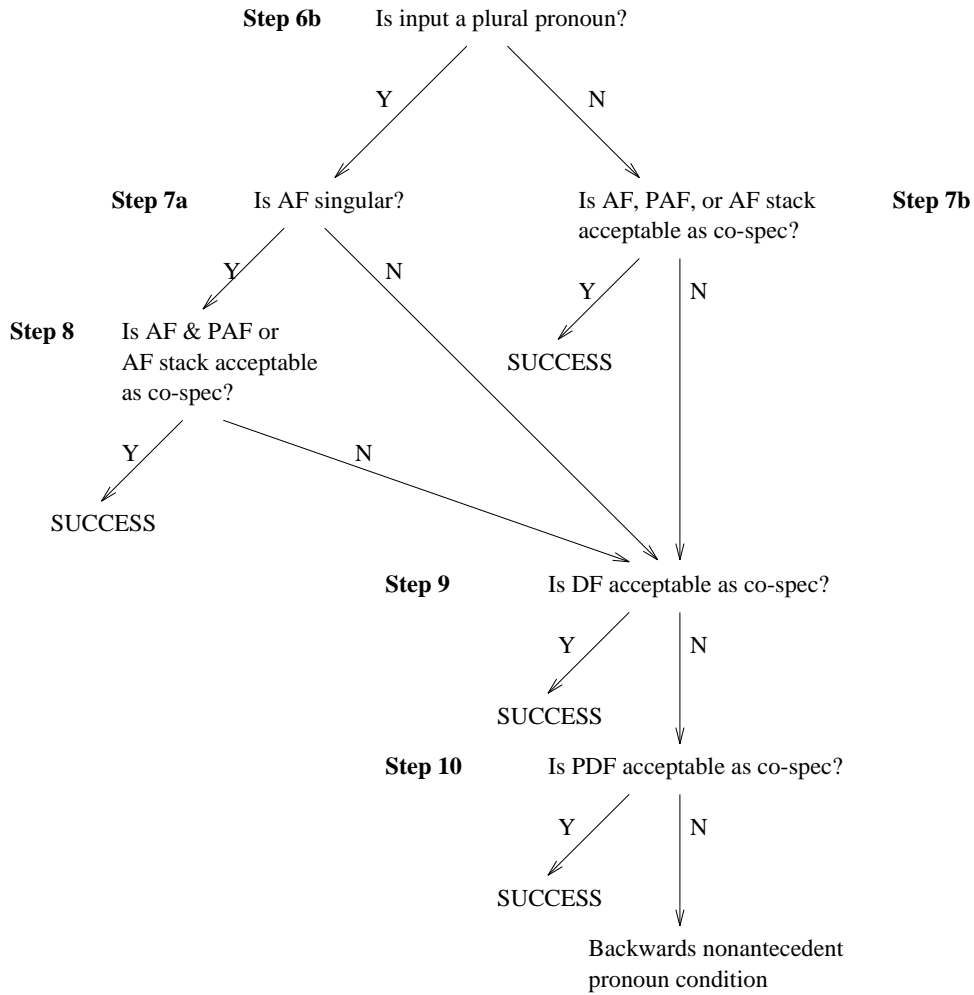

Figure 4.2: (Figure 4.1 cont.) Flowchart for the third person pronoun in agent position.

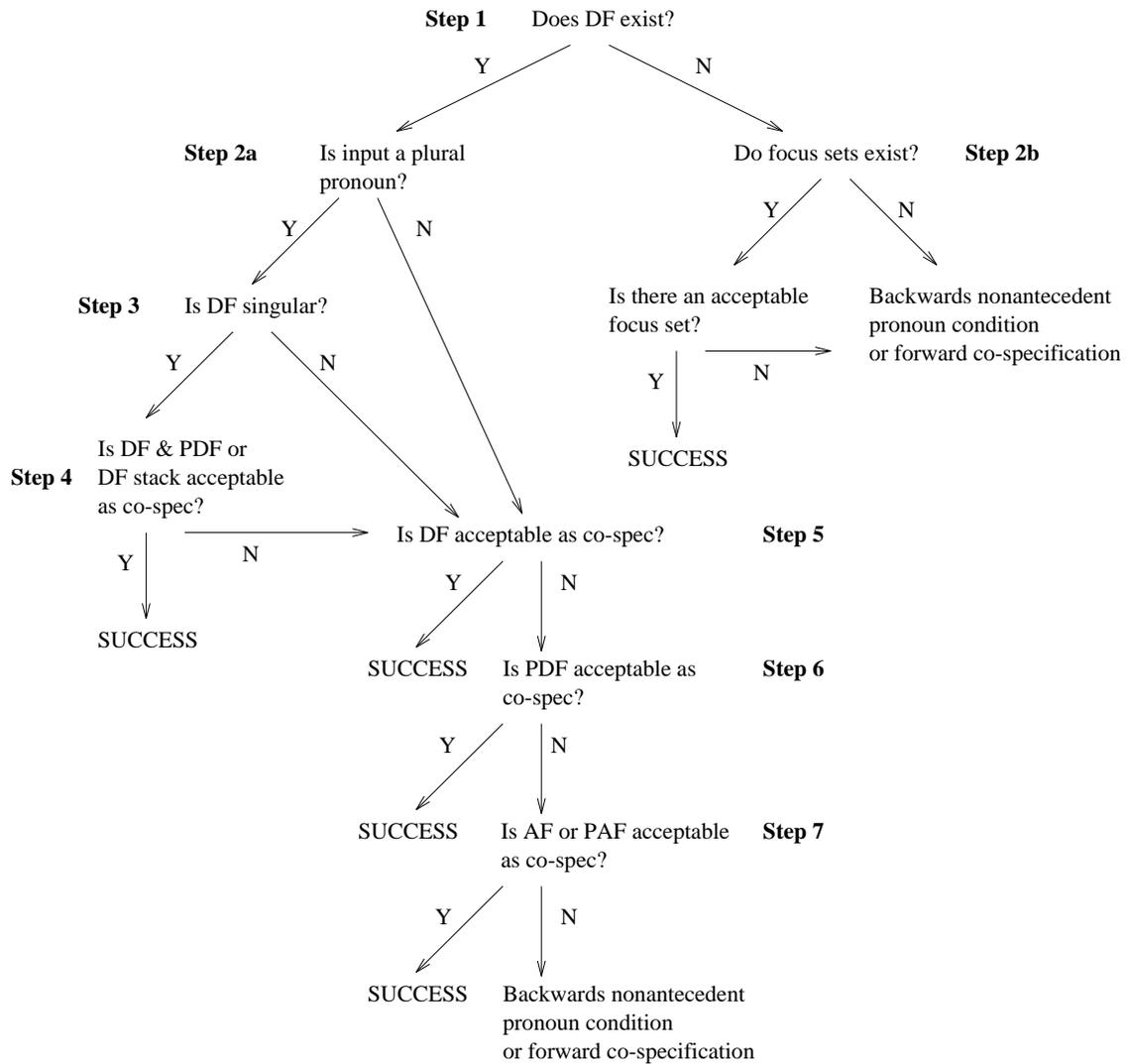

Figure 4.3: Flowchart for the third person pronoun in non-agent position.

The rules for the third person pronouns in non-agent position (cf. Figure 4.3) start with checking the existence of discourse focus. They are similar to the rules for the ones in agent position except that they do not check the actor focus and potential actor foci unless there is no other possibility. If there is no discourse focus, focus sets are checked in a manner similar to the above rules. If this check fails, *backwards non-antecedent pronoun condition or forward co-specification* holds. Forward co-specification condition is only possible for pronouns in non-agent position because of their nature. Again, similar to the above rules, according to the number of the pronoun under consideration DF, PDF, and DF stack are checked. If these checks do not end up with SUCCESS, AF and PAF are finally checked. If all of them fail, *backwards non-antecedent pronoun condition or forward co-specification* holds.

We have made an important modification in this part of the rules. In the original algorithm, Step 2 was performing the recency rule which reads "If the pronoun under consideration occurs in the *subject* position, ..." Therefore, it yields accurate results when applied to third person pronouns in agent position, not non-agent position. The rule is right, but not the algorithm. We have observed this fact during the experimentation phase and accordingly modified the algorithm. Sidner admits that the recency rule is problematic; it is not even clear that it should rule out focusing [47].

## 4.2 Evaluation and Analysis of the Experiments

### 4.2.1 Statistical Results

The statistical results of the program are presented in tabular form in Figure 4.4. The pronouns that are tried to be resolved by this program are listed on the left handside of the table. The pronoun *it* is presented in two separate rows, one representing the pronoun in agent position (**It(A)**), and the other in non-agent position (**It**). The reasons for this discrimination will be apparent when the columns are explained. The first column, **OCC**, holds the number of occurrences of the pronoun to its left. The second column, **RES**, indicates the number of occurrences which are successfully resolved. The last row sums up the previous rows and displays the overall results. It is seen from the first two columns of that row that the program has resolved 79 of 102 pronouns accurately, yielding a 77% success rate. The remaining nine columns contain the explanations of the remaining 23% and the effects of modifications to the original algorithms.

The third, fourth, and fifth columns are about the recency rule. The column **MRR** shows the effect of modifying the algorithm in the way that the recency rule only takes place in the algorithm for third person pronouns in agent position. The reasons of this modification were explained in the Section 4.1.2. As it is seen from the table, nine of the 11 occurrences of pronoun *it* in non-agent position are resolved by the program. Seven of them could be resolved accurately only after this modification has been made. Obviously, this modification only effects the pronouns in non-agent position with an acceptable object as the last noun phrase of the previous sentence. The next two columns **RR+** and **RR−** indicate the cases in which the recency rule (valid only for pronouns in agent position) caused the correct interpretation of the pronoun and in which it led to a wrong interpretation, respectively. Five cases would have been resolved inaccurately had the recency rule not been applied. On the other hand, five pronouns are misinterpreted because of this rule. These results show that the recency rule, even when it is applied only to cases in which the pronoun appears in agent position, is still problematic.

The sixth column, **INF**, indicates the cases in which the inferencing would provide resolving the pronoun accurately. There are three such cases. The necessity of inferencing (the ability to use world knowledge) is an unarguable fact appearing in the results of the most of the studies, and is also supported by our experimental results. The seventh column, **PA**, and the eighth column, **MPA**, are related. The former holds the number of cases in which *the potential actor ambiguity condition* has occurred in spite of the modification we have made. The latter holds the number of cases in which the modification helped resolve the pronoun accurately instead of marking the case as ambiguous. This modification was explained in Section 4.1.2. Seven potential actor ambiguity conditions are met and three of them are resolved by the help of that modification. The next column **BNFC** indicates the cases where the *backwards non-antecedent pronoun condition or forward co-specification* has occurred. There are three such cases. The numbers in this column are marked by a plus sign because we do not expect the algorithm to resolve the pronouns appearing in these discourse segments. In two of the cases, the pronoun has occurred in the first sentence and had an antecedent outside the discourse segment. One case is a representative of the special cases where focusing does not bring an explanation (examined in Section 2.3). These are the foreseen cases; *backwards non-antecedent pronoun condition or forward co-specification* is designed for marking this kind of cases. So, the algorithm should not be regarded as having failed actually. The tenth column, **BN**, indicates the case in which only the *backwards non-antecedent pronoun condition* has occurred. There is only one such case. In this case, the pronoun *it* is in agent position and its a sentential pronoun, i.e., it co-specifies with the verb phrase of the previous sentence.

The last column, **FS**, indicates the cases in which the pronoun cannot be resolved accurately and the real antecedent of the pronoun was in the focus stack but could not be reached by the program. There are six such cases. The use of a focus stack is an important and useful feature of focusing. The speakers can return back to a previously mentioned entity, discarding the intervening foci. Focusing is claimed to be able to model this process as explained in Section 4.1.2. The results show that it cannot, in most of the cases. The reason is that the focus stack is checked after the discourse focus, and potential foci. In our examples, the discourse focus or potential foci were acceptable as the co-specifications of the pronouns under consideration so, they are taken as the antecedents of pronouns. In some of the cases, the real antecedent was at the bottom of the stack. Even if the stack were checked there would be other entities to be checked first. It can be said that the algorithms cannot capture the focus popping process fully.

|        | OCC | RES | MRR | RR+ | RR- | INF | PA | MPA | BNFC | BN | FS |
|--------|-----|-----|-----|-----|-----|-----|----|----|------|----|----|
| **I**    | 20  | 20  |     |     |     |     |    |    |      |    |    |
| **You**  | 0   |     |     |     |     |     |    |    |      |    |    |
| **He**   | 19  | 14  |     | 1   |     |     | 2  | 3  |      |    | 3  |
| **She**  | 4   | 4   |     |     |     |     |    |    |      |    |    |
| **It (A)** | 16 | 10 |     | 5   | 5   |     |    |    |      | 1  |    |
| **It**   | 11  | 9   | 7   |     |     | 1   |    |    |      |    | 1  |
| **We**   | 6   | 2   |     |     |     | 2   |    |    | 2+   |    |    |
| **They** | 13  | 9   |     |     | 1   |     | 2  |    |      |    | 1  |
| **Me**   | 5   | 5   |     |     |     |     |    |    |      |    |    |
| **Him**  | 2   | 1   |     |     |     |     |    |    |      |    | 1  |
| **Her**  | 3   | 3   |     |     |     |     |    |    |      |    |    |
| **Us**   | 0   |     |     |     |     |     |    |    |      |    |    |
| **Them** | 3   | 2   |     |     |     |     |    |    | 1+   |    |    |
| **Total**| 102 | 79  | 7   | 6   | 6   | 3   | 4  | 3  | 3    | 1  | 6  |

Figure 4.4: Statistical results of the program.

### 4.2.2 A Closer Look at Some of the Examples

In this section, five different examples will be scrutinized. These exhibit different powerful features, or the deficiencies of the algorithms. Thirty-five discourse segments are used in the experimentation phase. Eleven of these discourse segments are taken from Sidner's thesis [48]. Six of them are among Hobbs' examples [23, 24]. In fact, three of them are well-known examples examined by several researchers other than Hobbs [23]. Three of the six are taken from the examples on which Hobbs' algorithm did not work [24]. Our program successfully resolved five pronouns occurring in these three examples. Eighteen of the discourse segments are taken from the articles published in *National Geographic* [39, 40]. The full set of discourse segments used in the experiments is reproduced in Appendix A.

**Example 1**

The first example is from Sidner's thesis [48, p. 86]. It represents a good example for the relation between anaphora usage and focusing process. In this example, there are two similar discourse segments which only differ in one sentence. This difference marks an important feature.

> 1 I got a new hat
> 2 and I decorated *it* with a big red bow.
> 3 a I think the bow will brighten *it* up a lot.
>   b I think *it* will brighten up the hat a lot.
> 4 If not, I guess I'll use *it* anyway.

The pronoun *it* in the second sentence no doubt co-specifies with *a new hat*. The interesting thing happens in the sentences 3 and 4. In 3a, *a big red bow* is referred to by a definite noun phrase, and *a new hat* by a pronoun. In 3b, it is the other way around; *a big red bow* is referred to by a pronoun, and *a new hat* by a definite noun phrase. If sentence 2 is followed by 3a, the pronoun *it* in sentence 4 tends to co-specify with the hat. If 3b follows sentence 2, the pronoun tends to co-specify with the bow. The discourse focus is the hat for 3a, and the bow for 3b. It is observed that there is a slight preference that the focussed entity will be referred to by a pronoun rather than a definite noun phrase. This is a natural consequence. Since pronouns contain less information than definite noun phrases, they are used to refer to the most salient entity in the discourse, the focus. This observation is reflected as a rule in Step 3 of the focusing algorithm.

**Example 2**

This is also from Sidner [48, p. 59]. In this example, it is understood that it is not possible to construct a system which works 100% accurately without inferencing (using the discourse and world knowledge). Inferencing is not only necessary for pronoun resolution, it is a must for discourse understanding.

> 1 I wanted to go to the movies on Saturday.
> 2 John would come too
> 3 but Bill decided to stay home.
> 4 So *we* went
> 5 and [*we*][7] afterwards had a beer.

---

[7]The pronoun *we* in the fifth sentence is missing in the original discourse but is syntactically recoverable. Therefore, it is represented in the data structure.

Sidner states that the discourse focus is necessary to decide who went to the movies among the three actors *I*, *John*, and *Bill*. However, focusing cannot establish the discourse focus correctly in this example; it cannot be aware of the difference between the sentences 2 and 3 without inferencing. The correct antecedent of the pronoun *we* is the set of *I* and *John* but the program finds the co-specification of *we* as the set of *I*, *John*, and *Bill* from the actor focus stack.

**Example 3**

The third example is a well-known one, examined by several researchers including Hobbs [23] and Winograd [59]. In fact, it consists of two discourse segments which are slightly modified here [48, p. 74]:

  a The city council refused to give the women a permit
    because *they* feared violence.
  b The city council refused to give the women a permit
    because *they* advocated revolution.

The antecedent of the pronoun *they* in sentence a is *the city council*. In sentence b, the antecedent is *the women*. The difference arises from the semantics of the sentences which cannot be captured by the focusing. However, focusing has some advantages over other techniques. It is widely accepted that inferencing is necessary to correctly interpret these sentences. Because focusing uses inferencing as a confirming mechanism, when the inferencing is complex, focusing becomes advantageous. Sidner states that "Focusing simplifies the inference process because it indicates what the beginning and endpoints of the inferencing are, and which inference can be taken back if a contradiction results" [48, p. 74]. For sentence b, Sidner claims that her algorithms will predict incorrectly that *they* co-specifies with *the city council*. During the inferencing process, the condition in which the city council both advocate the revolution and refuse to give a permit will be reached. At this stage, other schemes might look for another event to inference on, but focusing retracts from this point and chooses *the women* as the antecedent. The inferencing mechanism does not reject *the women* as the antecedent. However, when this discourse segment is given to the program, it outputs that *the potential actor ambiguity condition* is met. The program is correct; there are one actor focus (*the city council*) and one potential actor focus (*the women*). So, even if the inferencing is modeled computationally, the algorithms will not be able to resolve the pronoun correctly for this example.

**Example 4**

This is taken from Hobbs [24, p. 15]. He has excerpted this discourse segment from the novel *Wheels* by Arthur Hailey.

  1 The executive vice president had already breakfasted alone.
  2 A housekeeper had brought a tray to his desk.
  3 He had been alternately reading memoranda
  4 and [he had been] dictating crisp instructions into a recording machine.
  5 He had scarcely looked up.

Hobbs' algorithm picks the housekeeper as the antecedent of the pronoun *he* in sentence 5 and of *his* in sentence 2. It chooses *his* as the antecedent of *he* in sentence 3 and 4. If *his* were resolved correctly, this interpretation would be correct too. Because Hobbs' algorithm is designed for pronouns including possessives, it tries to resolve the pronoun *his*. In our system, this information is given manually to the system. Our program resolves all the pronouns in this example accurately without taking the housekeeper into consideration. After the first sentence is processed, the expected focus is the executive vice president. It is confirmed to be the discourse focus in the second sentence because *his desk* implicitly specifies it. Since the focus of the discourse is the executive vice president, it is checked first as the antecedent of the pronouns, and gets accepted. There is no need to check the housekeeper; the antecedent has been found. The representations of some of the data structures used in this example are shown in Figures 4.5–4.9. Figure 4.5 represents the hierarchical structure of the discourse segment. There are five sentences, 10 noun phrases, and five verb phrases. In Figure 4.6, the unit *sentence3* which represents sentence 3 is displayed. Because it has many slots, the whole unit cannot fit into a single window and a scroll bar is attached to it by KEE. In the next two figures (Figures 4.7-4.8), the units that represent the two noun phrases *his desk* and *he* are displayed. Some of the slots of units and their values can be seen in these figures. The noun phrase *his desk* is marked as a definite noun phrase and as the implicit specification of the noun phrase *the executive vice president* which is represented by the unit *np1*. The other noun phrase represented by unit *np7* is marked as a pronoun. This snapshot is taken after the execution of the program so the slot *co-spec* of this unit has the value unit *np1*. This slot had the value 'unknown' before the execution of the program. Figure 4.9 displays the unit *vp4* which represents the verb phrase of sentence 4.

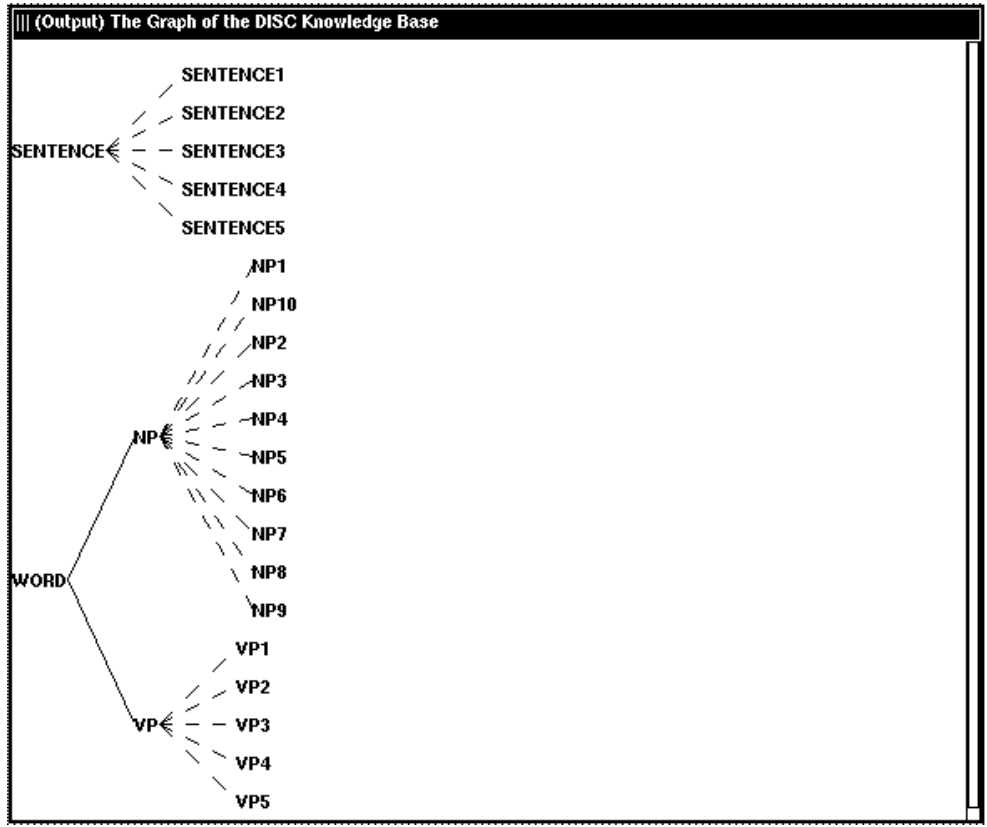

Figure 4.5: The hierarchical structure of the discourse segment in Example 4.

**Example 5**

This is taken from *National Geographic* [39, p. 21].

1 Ümit asked a man for the directions.
2 The talking stopped.
3 Men eyed him suspiciously.
4 He assured them,
5 he was a family friend.

The program accurately finds the antecedent of *them* as *men*. However, the antecedents of all the other pronouns are found as *a man*, whereas the correct antecedent is *Ümit*. Here the problem arises from the order in which the structures related to focus are checked. The noun phrase *a man* is a potential actor focus whereas *Ümit* is a member of actor focus stack. Because the potential actor foci are checked before the actor focus stack, *a man* is taken to be the co-specification. There is no syntactical, semantical, or inferencing clue to prevent this interpretation. If *Ümit* would have asked a woman instead of a man, the pronouns would be resolved accurately. This is a sort of rule called *the stacked focus constraint* by Sidner [48, pp. 88–89]: if an object from the focus stack is to be referred to by a pronoun, there should not be any other acceptable possibility in the focus or potential foci. We see that this rule can be overridden in some cases by human speakers.

```
||| (Output) The SENTENCE3 Unit in DISC Knowledge Base
Unit: SENTENCE3 in knowledge base DISC
Created by ebru on 6-20-94 21:02:39
Modified by ebru on 7-11-94 11:41:09
  Member Of: SENTENCE

Own slot: COMPLETE from SENTENCE3
 Inheritance: OVERRIDE.VALUES
 ValueClass: INTEGER
 DefaultValue: 1
 Comment: "This slot is 0 if the sentence is not complete, i.e., the following sentence is tied to that sentence."
 Values: 0

Own slot: DEFINITE-ANAPHORA from SENTENCE3
 Inheritance: OVERRIDE.VALUES
 ValueClass:
         (LIST.OF (ONE.OF SUBJECT NP1 NP2))
 Comment: "The position of anaphoric expressions in the sentence"
 Values:
       (SUBJECT)

Own slot: NEXT-SENTENCE from SENTENCE3
 Inheritance: OVERRIDE.VALUES
 ValueClass:
         (MEMBER.OF SENTENCE)
 Values: SENTENCE4

Own slot: NP1 from SENTENCE3
 Inheritance: OVERRIDE.VALUES
 ValueClass:
         (LIST.OF (MEMBER.OF NP))
 Values:
       (NP6)

Own slot: NP2 from SENTENCE
 Inheritance: OVERRIDE.VALUES
 ValueClass:
         (LIST.OF (MEMBER.OF NP))
 Values: UNKNOWN

Own slot: NTYPE from SENTENCE
 Inheritance: OVERRIDE.VALUES
 ValueClass:
         (ONE.OF IS-A THERE-INSERTION CLEFT PSEUDO-CLEFT NORMAL)
 DefaultValue: NORMAL
 Cardinality.Max: 1
 Cardinality.Min: 1
 Values (defaulted): NORMAL
```

Figure 4.6: The unit which represents sentence 3 in Example 4.

```
||| (Output) The NP4 Unit in DISC Knowledge Base
Unit: NP4 in knowledge base DISC
Created by ebru on 4-23-94 14:59:21
Modified by ebru on 7-3-94 14:18:24
Member Of: NP

Own slot: ANAPHOR from NP4
Inheritance: OVERRIDE.VALUES
ValueClass:
        (ONE.OF DEFNP PRONOUN)
Values: DEFNP

Own slot: BELONGS-TO from NP4
Inheritance: OVERRIDE.VALUES
ValueClass:
        (MEMBER.OF SENTENCE)
Values: SENTENCE2

Own slot: CO-SPEC from NP4
Inheritance: OVERRIDE.VALUES
ValueClass:
        (LIST.OF (MEMBER.OF NP))
Values:
    (NP1)

Own slot: GENDER from NP4
Inheritance: OVERRIDE.VALUES
ValueClass:
        (LIST.OF (ONE.OF M F N))
Values:
    (N)

Own slot: IMPLICIT-SPEC from NP4
Inheritance: OVERRIDE.VALUES
ValueClass: INTEGER
Values: 1

Own slot: LIFE from NP4
Inheritance: OVERRIDE.VALUES
ValueClass:
        (ONE.OF ANIMATE INANIMATE)
Values: INANIMATE

Own slot: NUMBER from NP4
Inheritance: OVERRIDE.VALUES
ValueClass:
        (ONE.OF SINGULAR PLURAL)
Values: SINGULAR
```

Figure 4.7: The unit which represents the noun phrase *his desk* in sentence 2 in Example 4.

```
||| (Output) The NP7 Unit in DISC Knowledge Base
Unit: NP7 in knowledge base DISC
Created by ebru on 5-12-94 14:24:43
Modified by ebru on 7-11-94 11:52:16
Member Of: NP

Own slot: ANAPHOR from NP7
Inheritance: OVERRIDE.VALUES
ValueClass:
        (ONE.OF DEFNP PRONOUN)
Values: PRONOUN

Own slot: BELONGS-TO from NP7
Inheritance: OVERRIDE.VALUES
ValueClass:
        (MEMBER.OF SENTENCE)
Values: SENTENCE4

Own slot: CO-SPEC from NP7
Inheritance: OVERRIDE.VALUES
ValueClass:
        (LIST.OF (MEMBER.OF NP))
Values:
     (NP1)

Own slot: GENDER from NP7
Inheritance: OVERRIDE.VALUES
ValueClass:
        (LIST.OF (ONE.OF M F N))
Values:
     (M)

Own slot: IMPLICIT-SPEC from NP
Inheritance: OVERRIDE.VALUES
ValueClass: INTEGER
Values: 0

Own slot: LIFE from NP7
Inheritance: OVERRIDE.VALUES
ValueClass:
        (ONE.OF ANIMATE INANIMATE)
Values: ANIMATE

Own slot: NUMBER from NP7
Inheritance: OVERRIDE.VALUES
ValueClass:
        (ONE.OF SINGULAR PLURAL)
Values: SINGULAR
```

Figure 4.8: The unit which represents the noun phrase *he* in sentence 4 in Example 4.

```
||| (Output) The VP4 Unit in DISC Knowledge Base
Unit: VP4 in knowledge base DISC
Created by ebzu on 5-17-94 9:45:57
Modified by ebzu on 7-3-94 14:13:43
 Member Of: VP

Own slot: BELONGS-TO from VP4
 Inheritance: OVERRIDE.VALUES
 ValueClass:
         (MEMBER.OF SENTENCE)
 Values: SENTENCE4

Own slot: TEXT from VP4
 Inheritance: OVERRIDE.VALUES
 ValueClass: STRING
 Comment: "The word itself."
 Values: "HAD-BEEN-DICTATING"

Own slot: THEME from VP4
 Inheritance: OVERRIDE.VALUES
 ValueClass:
         (LIST.OF (MEMBER.OF NP))
 Values:
       (NP8)
```

Figure 4.9: The unit which represents the verb phrase *had been dictating* of sentence 4 in Example 4.

# Chapter 5

# Conclusion and Future Work

In this study, focusing algorithms devised by Sidner [48] were implemented and evaluated. Focusing phenomenon was first examined by Grosz [17] and Sidner. Sidner devised several algorithms to model the focusing process and to resolve anaphora using focusing as a tool. These studies on focusing form a basis for some of the recent studies. Sidner's algorithms have only partial implementations. A complete implementation is needed to examine the algorithms from a computational point of view, and to be able to conduct experiments for testing the success of the algorithms.

Our experiments showed that although the algorithms have some deficiencies, the system is modestly successful overall (77%). It can be improved to give better results in the light of recent studies and with the help of relevant discourse information such as stress and intonation [6, 7]. Implementing an inference mechanism will also increase the success of the system. The theory can also be said to be computationally feasible. The simplicity of the rules makes them easily implementable. One major contribution is on the subject of inferencing. Because focusing reduces the job of inferencing to directly supporting a focus prediction rather than producing the co-specification of an anaphoric expression as a result of a general inferencing, it becomes computationally realizable.

There are three future research directions that can be taken. This study deals only with pronouns. It can be enlarged to cover other kinds of anaphora. There are four additional kinds of anaphora examined in Sidner's thesis. After implementing algorithms for these kinds of anaphora, new algorithms can be devised for the remaining kinds of anaphora such as generic anaphora, one anaphora, etc. The second direction of research is the use of focusing for language generation. In Sidner's thesis, focusing is examined from the standpoint of hearers, i.e., the process in which the hearers comprehend anaphoric expressions is studied. Another study can be conducted to understand the process in which the anaphoric expressions are generated. Another research direction, presently the most favored one by us, is to build a system which resolves anaphora in Turkish discourse. In addition to pronouns, different kinds of anaphora appearing in Turkish discourse can be examined and a system can be implemented to resolve them. The current studies on anaphora in Turkish use other techniques than focusing [53, 52]. The focusing process should be reformulated according to the syntactic features of Turkish sentences. The anaphora resolution algorithms should also be modified to reflect the anaphora usage in Turkish discourse.

# Appendix A

# Discourse Segments

The following discourse segments were used in the experimentation phase. The pronouns in bold face indicate the ones that could not be resolved by our program. All other pronouns were correctly resolved.

## A.1 Discourse segments taken from Sidner

I lost a necklace at the office yesterday.
I inherited it from my grandmother and it meant a lot to me.

I want to schedule a meeting with George, Jim, Steve, and Mike.
We can meet in my office.
It's kind of small, but the meeting won't last very long anyway.

I got a new hat and I decorated it with a big red bow.
I think the bow will brighten it up a lot.
If not, I guess I will use it anyway.

I got a new hat and I decorated it with a big red bow.
I think it will brighten up the hat a lot.
If not, I guess I will use it anyway.

John is an orphan.
**He** misses **them** very much.

John and Mary played cards.
Henry read a book.
They went to Joey's Bar to hear a new rock group.

Jerome took his pigeon out on a leash.
Since he was trying to train it, he hollered "heel" and "run" at it, as they sauntered along.

I wanted to go to the movies on Saturday.
John would come too but Bill decided to stay home.

So **we** went and **we** afterwards had a beer.

Last week, we went out to the lake near my cottage.
**It** was a lot of fun.

Wilbur is a fine scientist and a thoughtful guy.
He gave me a book.
It was on relativity theory.
**It** talks about quarks.
They are hard to imagine, because they indicate the need for elementary field theories.
These theories are absolutely essential to relativity research.
Anyway, I got **it**, while I was working on the initial part of my research.
**He** is really a helpful colleague.

I had a problem with my radio, because the speaker made a buzzing noise.
I decided to take it to be fixed.

## A.2   Discourse segments taken from Hobbs

The city council refused to give the women a permit because **they** feared violence.

The city council refused to give the women a permit because **they** advocated revolution.

Jack invited Janet to his birthday party.
She wondered, if he would like a kite.
But Mary said, he already had a kite.
He would make her take **it** back.

Should not someone answer?
If some bright reporter gets to Henry Ford, he is apt to.

The executive vice president had already breakfasted alone.
A housekeeper had brought a tray to his desk.
He had been alternately reading memoranda and he had been dictating crisp instructions into a recording machine.
He had scarcely looked up.

CITMOCO asked the Commerce Department for an export license, and CITMOCO got it within two days.

## A.3   Discourse segments taken from National Geographic

Ümit asked a man for the directions.
The talking stopped.
Men eyed **him** suspiciously.

**He** assured them, **he** was a family friend.

One hot afternoon I was sitting on a shady stone wall, when a gate opened behind me, and a similing man presented me with a cup of cool ayran.
Then another gate opened and his neighbor emerged with ayran.
**They** stood silently there.

We don't often hear about happiness from Native Americans.
For hundreds of years their story has been sad.
They have lost a great, free way of life.

Not finished yet, the 12-foot-long beluga whale backed away a little and blew the air out once more.
This time he nodded his head sharply downward, [he][1] sending an invisible boil of water against the expanding bubble.
It instantly became a twisting bracelet, shining and expanding until **it** began to break into flattened, rising spheres.

The belugas below us at Cunningham Inlet were yellowish white ghosts against the jade water.
They wriggled against one another like tadpoles.

Beluga milk can be eight times richer than cow's milk.
Capable of supporting the baby's rapid growth, **it** quickly provides the young one with a warm blanket of blubber.

Instead, the pesticide appears to be carried to the St. Lawrence by eels migrating from the Great Lakes, where it's still found in sediment.

I don't use tobacco.
It may stunt my growth.

Puzzled, I ask announcer Dennis Heggenstaller about the weight contraptions.
He looks at me as if I've just flown in from Jupiter.

As I enter the parking lot, a little blond sprite in a blue dress bounds from the plain, wooden store and barefoot, traverses walnut size gravel as if it is carpet.
At my fender she stops and raises her right arm.

Round-faced Emma has been in school only two weeks, but she is confident when Frona asks her to count to a hundred and [she][2] needs only one assist.

When the men awoke the next morning, they found themselves surrounded by 50-some Seneca and two British officers in pint, a guerilla band bent on terrorizing the frontier by collecting scalps for the clown.

---

[1] Syntactically recoverable object.
[2] Syntactically recoverable object.

At midday the heat was relentless.
Clouds of dust rose from the road and swirled into the bus as it rattled through the forest.

In the arena voices shouted through the public-address system.
**It** sounded as if three men were having a fight, but **it** was almost the opposite.
Powwows today are far more than a salute to the past.
They're not shows.
**They**'re not entertainment.
Most Indians call them celebrations.

Old Horn is a Crow from Montana.
He wears a black vest and black hat.
His black hair goes halfway down his back; today long hair is a symbol of Indian pride.
He's tall and lean as a knife.

Indians are practical; they use whatever comes to hand.

Jonathan is also singing with his drum group, Haystack, and I ask him if we can talk for a while about singing and the powwow.